\documentclass[letterpaper, 10 pt, journal, twoside]{IEEEtran}

\IEEEoverridecommandlockouts                              % This command is only needed if 
\usepackage{todonotes}
\usepackage{graphicx} % for pdf, bitmapped graphics files
\usepackage{mathptmx} % assumes new font selection scheme installed
\usepackage{amsmath}
\newcommand{\norm}[1]{\left\lVert#1\right\rVert}
\usepackage{bm}
\usepackage{mathtools} % assumes amsmath package installed
\usepackage{amssymb}  % assumes amsmath package installed
\usepackage{algorithm}
\usepackage{algorithmic}
\usepackage{subfigure}
\usepackage{url}
\graphicspath{{./fig/}}
\usepackage{tikz}
\usepackage{lipsum}

\newcommand\copyrighttext{%
	\footnotesize \textcopyright 2020 IEEE. Personal use of this material is permitted. Permission from IEEE must be obtained for all other uses, in any current or future media, including reprinting/republishing this material for advertising or promotional purposes, creating new collective works, for resale or redistribution to servers or lists, or reuse of any copyrighted component of this work in other works.}
\newcommand\copyrightnotice{%
	\begin{tikzpicture}[remember picture,overlay]
	\node[anchor=south,yshift=10pt] at (current page.south) {\fbox{\parbox{\dimexpr\textwidth-\fboxsep-\fboxrule\relax}{\copyrighttext}}};
	\end{tikzpicture}%
}
\begin{document}

\title{Force-guided High-precision Grasping Control of Fragile and Deformable Objects using sEMG-based Force Prediction}

% Make room for more info lines in the \author command

\author{Ruoshi Wen, Kai Yuan, Qiang Wang Shuai Heng, Zhibin Li%

%	\thanks{This work was supported by the National Natural Science Foundation	of China under Grant 61876054, the China Scholarship Council, the EPSRC CDT in Robotics and Autonomous Systems (EP/L016834/1), and EPSRC Future AI and Robotics for Space (EP/R026092/1).}%Use only for final RAL version
%
%	\thanks{$^{1}$Ruoshi Wen and Qiang Wang are with the Department of Control Science and Engineering, Harbin Institute of Technology, Harbin 150001, China
%		{\tt\small ruoshiwen@gmail.com, wangqiang@hit.edu.cn}}%
%
%	\thanks{$^{2} $Kai Yuan and Zhibin Li are with the School of Informatics, The University of Edinburgh, Edinburgh EH8 9BT, U.K.
%
%		{\tt\small kai.yuan@ed.ac.uk, zhibin.li@ed.ac.uk}}%
%
%	\thanks{$^{3}$Shuai Heng is with the State Key Laboratory of Robotics and System, Harbin Institute of Technology, Harbin 150080, China
%		{\tt\small heng13514479054@163.com}}

}

%Use only for final RAL version. 

\maketitle
% Use only for final RAL version
\copyrightnotice

%%%%%%%%%%%%%%%%%%%%%%%%%%%%%%%%%%%%%%%%%%%%%%%%%%%%%%%%%%%%%%%%%%%%%%%%%%%%%%%%
\begin{abstract}
Regulating contact forces with high precision is
crucial for grasping and manipulating fragile or deformable objects. We aim to utilize the dexterity of human hands to regulate the contact forces for robotic hands and exploit human sensory-motor synergies in a wearable and non-invasive way. We extracted force information from the electric activities of skeletal muscles during their voluntary contractions through surface electromyography (sEMG). We built a regression model based on a Neural Network to predict the gripping force from the preprocessed sEMG signals and achieved high accuracy ($R^2 = 0.982$). Based on the force command predicted from human muscles, we developed a force-guided control framework, where force control was realized via an admittance controller that tracked the predicted gripping force reference to grasp delicate and deformable objects. We demonstrated the effectiveness of the proposed method on a set of representative fragile and deformable objects from daily life, all of which were successfully grasped without any damage or deformation.
\end{abstract}

% Keywords appear just beneath the abstract. Use only for final RAL version.
%\begin{IEEEkeywords}
%Dexterous Manipulation, Grasping, Force Control
%\end{IEEEkeywords}
%%%%%%%%%%%%%%%%%%%%%%%%%%%%%%%%%%%%%%%%%%%%%%%%%%%%%%%%%%%%%%%%%%%%%%%%%%%%%%%%
\section{Introduction}
\IEEEPARstart{H}{umans} can grasp most delicate and soft objects thanks to their adaptability of grasping forces through learned motor skills. Compared to humans, robots have a relatively limited grasping ability. Numerous deformable and fragile objects with unknown shapes and material properties can be found in daily life, manufacturing processes, agriculture, and outer space. Since humans can grasp these objects without causing damage, we are motivated to transfer this adaptability skill to robots. 
\begin{figure}[t]
	\centering
	\subfigure[]{
	\label{fig:pick_up_robot}
	\includegraphics[trim={1cm 0.45cm 1cm 0.5cm},clip, height=0.170\textwidth]{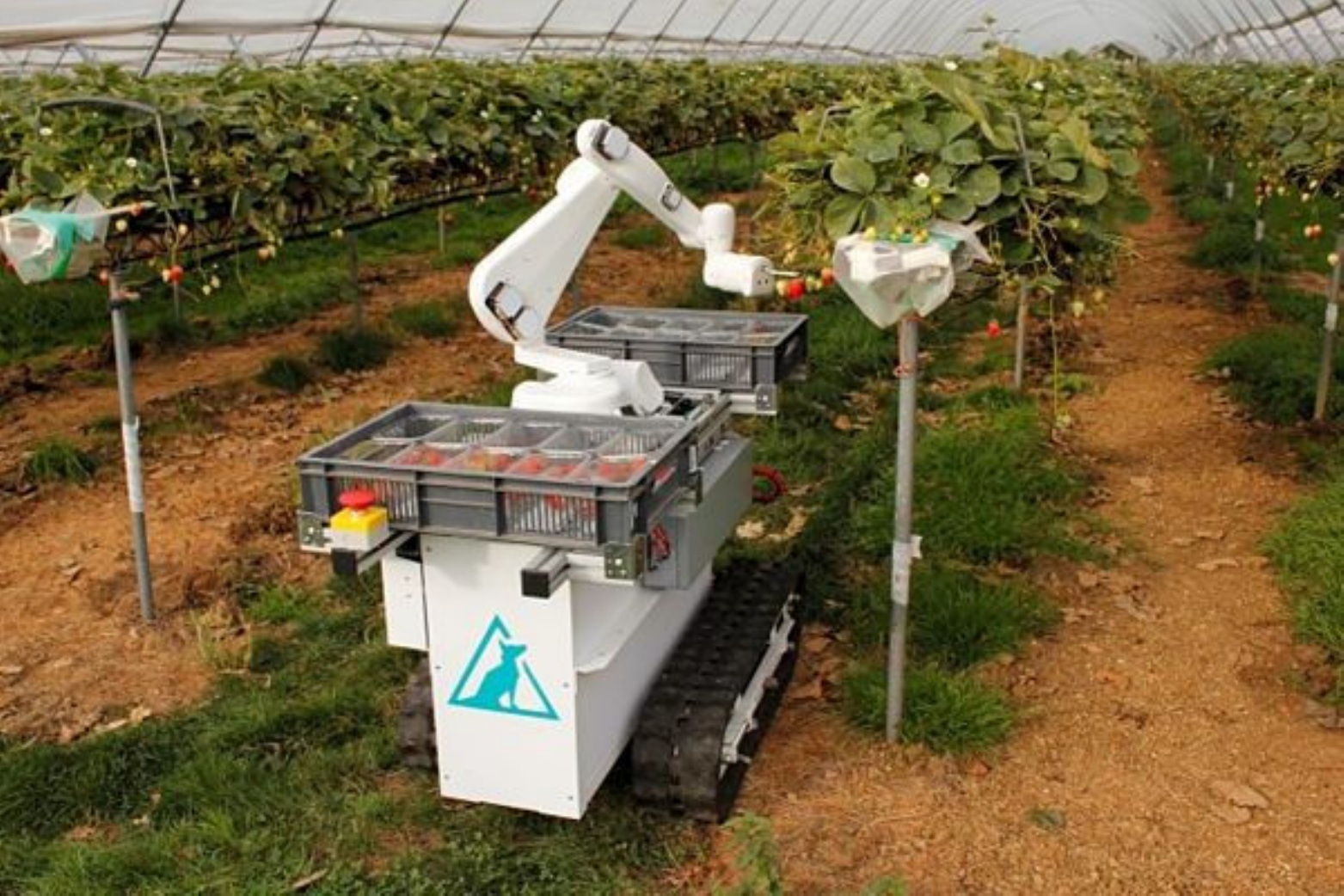}}
	\subfigure[]{
		\label{fig:straw}
		\includegraphics[trim={0.5cm 0.5cm 0cm 0.5cm},clip,height=0.17\textwidth]{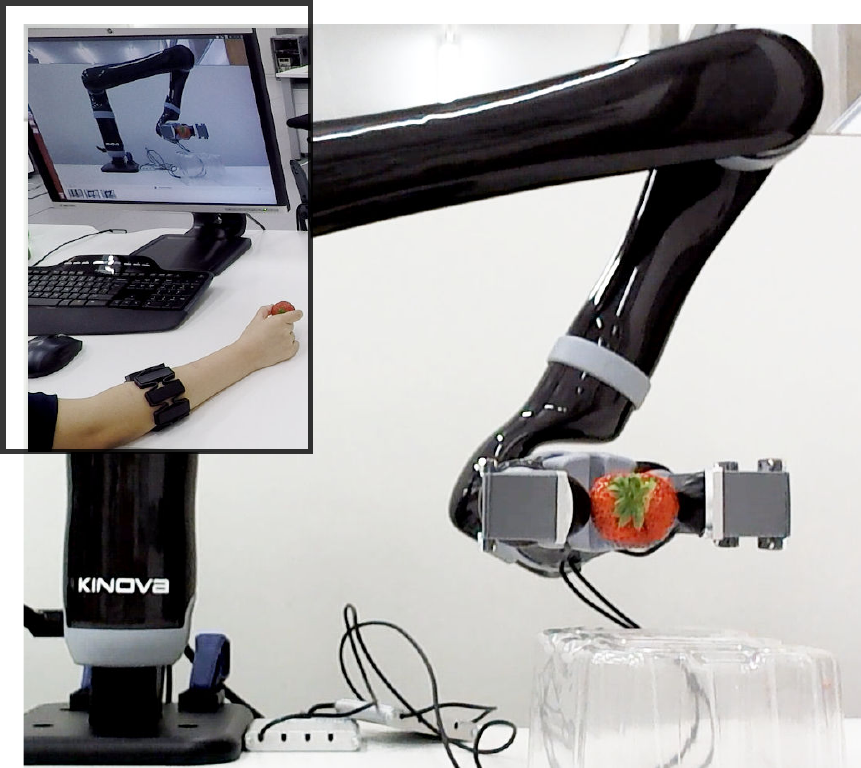}}
	\subfigure[]{
	\label{fig:objs}
	\includegraphics[trim={0cm 0cm 0cm 0cm},clip,height=0.184\textwidth]{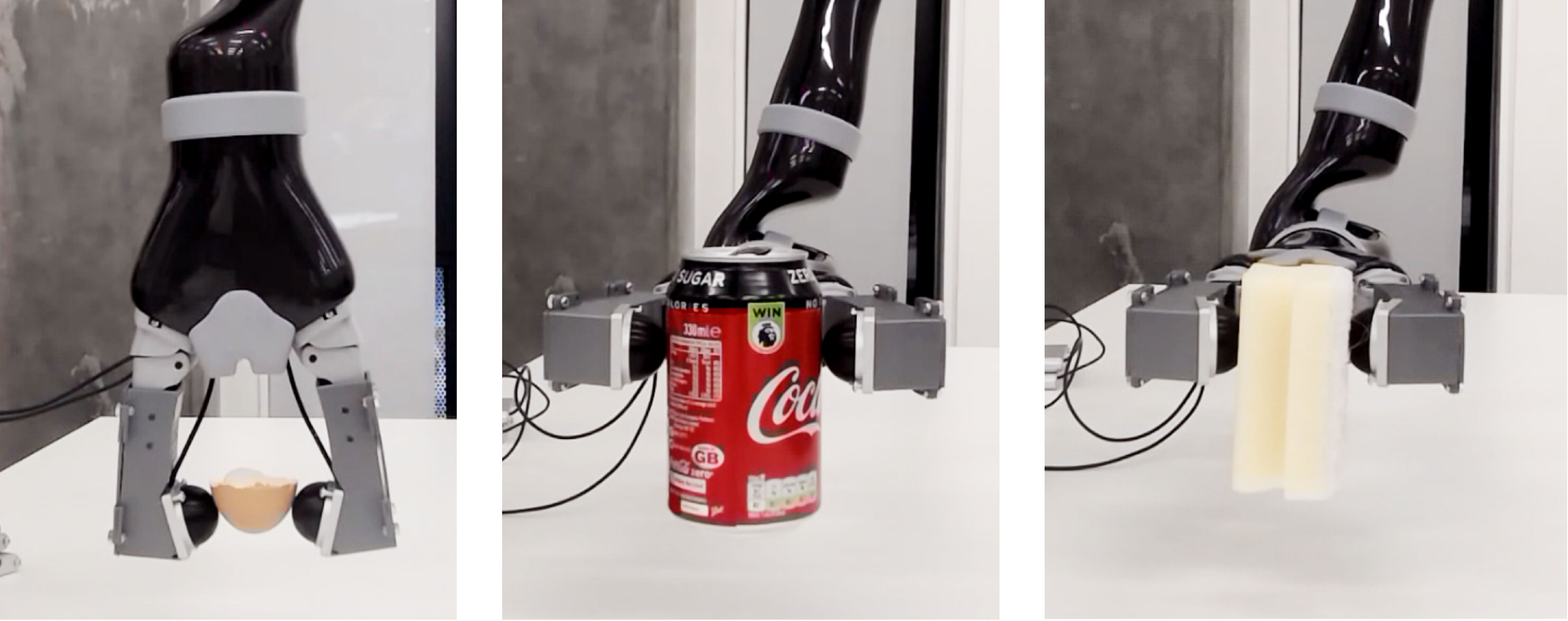}}
	
	\caption[]{Applications requiring delicate force controlled grasping. (a) an agriculture robot for harvesting strawberries (image courtesy: Dogtooth Technologies Limited); (b)-(c) grasping fragile objects via the proposed force-guided control system: a strawberry, an eggshell, an aluminum can and a sponge.}
	\vspace{-4mm}	
\end{figure}
To further extend robots' existing capability of manipulating rigid objects by adding the ability to grasp delicate and deformable objects, we aim to combine the humans' adaptability of grasping force with robots' ability to complete repetitive and dangerous tasks with high precision.

Given the global demographic skew towards an aging population, it is highly probable that there will be a requirement in the foreseeable future for service robots to assist in taking care of the elderly \cite{Portugal2019}. Among the emerging concerns, safety has the highest priority since robots will be involved in tasks such as handling delicate objects in proximity to humans and helping patients to move during rehabilitation. To address these concerns when designing service robotic systems, at the first stage of the development, the guidance/supervision from human demonstrators/operators can be used for strict contact force regulation to guarantee safety between humans and robots during interactions. 

Moreover, to assist forearm amputees with regaining their upper extremities' grasping ability (thus improving their life quality), researchers have been working on building advanced prosthetic devices. However, compared to humans' upper limbs, these prostheses currently cannot achieve the same level of precise, adaptable control over contact forces to handle a broad range of daily objects.  

For mass production in agriculture and industry, teleoperation techniques allow workers to be remotely involved in the manufacturing process. Humans' adaptation of grasping force for different tasks can improve the robots' manipulation ability. Fig. \ref{fig:pick_up_robot} shows that a robot is harvesting strawberries in a plantation and Fig. \ref{fig:straw} shows that a Kinova arm is teleoperated to pick up a ripe strawberry via a myoelectric interface (sEMG). Both provide ideas for future working scenarios where workers can get involved in agriculture remotely to pick up vegetables or fruits. Fig. \ref{fig:objs} shows that the gripper has successfully grasped representatives (a half eggshell, an empty aluminum can, and a sponge) of extremely fragile and deformable objects, indicating that the guidance from humans can help industrial robots to grasp products made of very fragile or deformable materials.
	
In the area of space robotics, shared autonomy between humans and robots plays an essential role in achieving Level E1 autonomy for onboard spacecrafts defined by European Cooperation for Space Standardization (ECSS) \cite{gao2016space}. Teleoperated robots can act as human \textit{avatars} working in environments that would be too dangerous or strenuous for humans. Notable examples include executing tasks in outer space, moving dangerous objects (e.g., explosive, nuclear, or toxic materials), and rescuing victims in disaster response situations. In these scenarios, precise contact force control for grasping and manipulating deformable and fragile objects is crucial due to the high costs of human life or equipment in case of failure. 

Motivated by the unresolved problems in the aforementioned areas, in this paper, we propose a force-guided high-precision control framework for fragile and deformable objects. While a human demonstrator is grasping objects, the contact forces are predicted via a wearable, non-invasive myoelectric interface realized by surface electromyography (sEMG) and transmitted to a robot's controller as force references. Within its payload, a robot can grasp and manipulate the same objects that a demonstrator can do.

The significance of this study is the development of an effective force-guided control interface for shared telemanipulation. Since there still remains substantial research to do before realizing fully dexterous autonomous grasping, this work provides human-in-the-loop solutions for safety-critical grasping tasks requiring fine force control. Moreover, this scheme and interface provide attractive ideas for transitional technologies towards autonomous grasping: to build large data sets of contact forces, spatial motions, and commands from humans. All these data capture the sensory-motor synergies, particularly high dimensional representations of motion and force primitives that are critical for generating multi-contact exploration and reactive behaviors. These useful data from demonstrations can facilitate future researches on learning-based autonomous grasping, e.g., training deep neural networks in a supervised manner to speed up the learning process. In addition, our method can be applied to most robotic arms, and upgrade the existing systems at a minimal hardware cost.

To build a gripping force prediction model, force sensors were used to gather data which were fed to a supervised neural network as labels. To realize force control on the robot, the compliant contact surface between the rubber-coated gripper and objects were modeled as a spring-mass-damper system, where an admittance controller was designed to track the predicted reference gripping forces.

The contributions of this work are summarized as follows:
\begin{itemize}
	\item A force-guided control framework for grasping fragile and deformable objects.
	\item Design of a {real-time, online} regression model from sEMG signals to reference gripping forces.
	\item Handling noise from sEMG signals through a combination of signal processing and the intrinsic properties of Neural Networks.
	\item Portable, wearable and low-cost Myo armband setup with high prediction accuracy ($R^2 = 0.982$).
	\item A hardware-independent framework for most robot platforms' development or existing systems' upgrade. 
\end{itemize}

The letter is organized as follows. Section \ref{sec:ii} summarizes the related works. In Section \ref{sec:iii}, we overview the force-guided grasping control framework and describe admittance control for force tracking. In Section \ref{sec:iv}, we present details of the prediction model and its real-time application. In Section \ref{sec:experiment setup}, we explain the setup of sensors for data acquisition and the robotic gripper for grasping tasks. In Section \ref{sec:results}, we first evaluate the performance of the prediction model, then further evaluate the grasping performance. In Section \ref{sec:discuss}, we discuss the reasons for the success and the generalization on other robot systems. We also analyze the limitations of this work and investigate their potential solutions. In the last section, we draw conclusions and propose some future works.

\begin{figure*}[t!]
	\centering
	\includegraphics[trim={0cm 0.3cm 0cm 0.1cm},clip,width=0.9\textwidth]{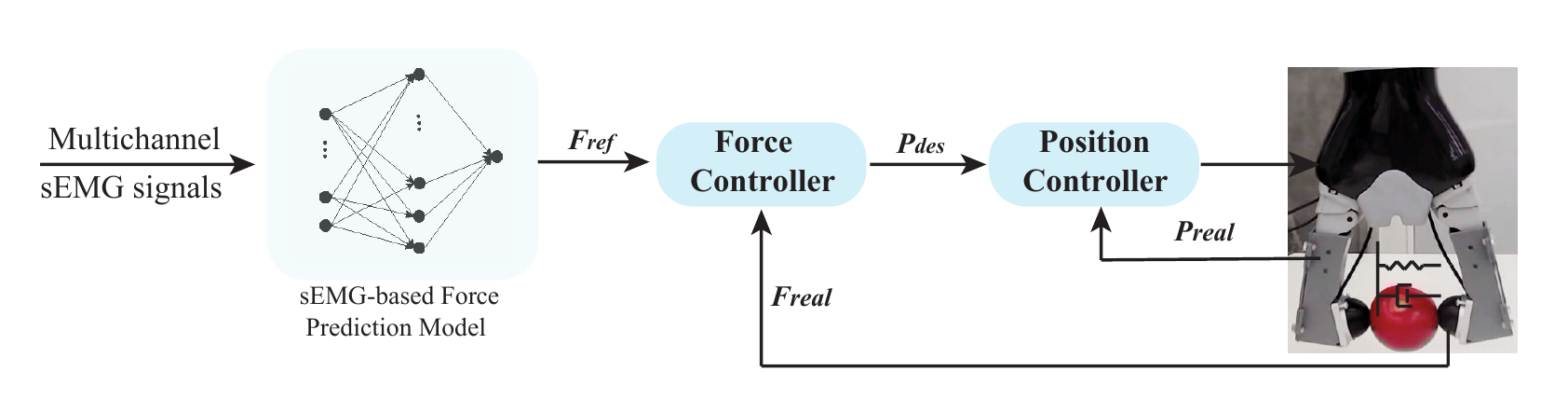}
	\caption{Force-guided grasping control framework with $F_{ref}$ predicted from the multichannel sEMG signals.}
	\label{fig:control_diag}
	\vspace{-6mm}
\end{figure*}

\section{Related Work}
\label{sec:ii}
Fragile and deformable objects' grasping and manipulation need fine-tuned force control which cannot be achieved by position controllers or traditional rigid end-effectors. In order to control robots to grasp these objects while guaranteeing safety, a trade-off between maintaining intact shape and avoiding slippage requires regulating the grasping force through sophisticated controllers. Researchers have developed slip and deformation (fracture) detection models with the aid of tactile sensors \cite{Zaidi2017}. The slip detection method needs the prior knowledge of an object such as weight and involves slip trials to estimate the friction coefficient of the contact surface. To reduce deformation, an upper bound of the normal force is also required  \cite{Kaboli2016a}, which is obtained through many trials or set empirically and cannot be set to the same value for all objects. The requirements and constraints make it difficult and time costly to apply these detection methods to unknown objects. From a  bio-mimic perspective, soft actuator techniques (fluid fingertips \cite{Nishimura2019}, pneumatic humanoid hand \cite{Salem2019}) have also been introduced for these tasks. Soft actuators are fabricated from flexible, soft and light materials. Compared to rigid end-effectors, the softness and compliance of these materials allow robots to interact safely with objects. The inherent properties of soft actuators reduce the control complexity and are advantageous for unknown objects, even with odd geometries or unusual surfaces \cite{Shintake2018}. Due to their pneumatic or fluidic driven methods, the challenges lie in slow response, theoretical modeling, insufficient strength, system integration, and miniaturization.

A human's central nervous system (CNS) is capable of a fast trial-by-trial adaptation to changes in a manipulation context, e.g., object weight, surface friction, and weight distribution \cite{Watanabe:2018:HID:3285138}. Demonstration-based robotic control can integrate humans' sensory-motor memory and decision intelligence into robots. For force-guided grasping control, it is crucial to get the control strategy from human demonstrators, such as the estimation of stiffness for variable impedance learning \cite{AbuDakka2018}. Computer vision (CV) techniques have been applied to get motion information such as joint positions, moving trajectories and interaction forces \cite{Pham2018, Wang2017}. In addition, haptic feedback provides a solution for obtaining the demonstrators' stiffness and contact force. In \cite{Kronander2014}, an interface for teaching compliance variations via haptic feedback is presented. In \cite{Li2014}, the stiffness was modulated online through human-robot interaction to achieve object-level impedance control for dexterous manipulation. 

The movements of the joints are driven by skeletal muscles, which work as actuators and receive the control commands from CNS. The contact force measured by haptic devices or estimated by CV techniques between the fingertips and objects is the execution result of skeletal muscles. Therefore, for real-time applications, predicting the movement information from skeletal muscles can compensate for the electromechanical delay (EMD) between the human and robot side. Moreover, haptic gloves constrain the fingers' moving range and lose the periphery feedback from skins, thus influencing humans' decision process. CV-based contact force estimation technique limits humans' activity space due to cameras' effective range. It also requires structural environments and the aid of markers to achieve better performance, which is not ideal in such applications, e.g., estimating contact forces to control prosthetic hands, teleoperating robots in an unknown environment such as in outer space. Estimating the contact force from skeletal muscles can free human demonstrators' hand movements, and does not put limitations on their activity range. 

Surface electromyography (sEMG) can be used to predict the contact force from skeletal muscles, which is a non-invasive procedure that involves the detection, recording, and interpretation of myoelectric activities \cite{c4}. It has wide applications in robotics such as advanced prostheses \cite{Gibson2015} and exoskeletons \cite{Artemiadis2009}. In \cite{Haas2018}, under conditions of severely impaired feedback, a robotic hand achieved variable-stiffness grasps of small-fruit containers, where the stiffness was acquired via sEMG from hand muscles. Compared to the task of grasping containers, our work includes a wider range of common delicate and deformable objects, which require fine-tuned force controllers. 
\begin{figure}[h!]
	\centering
	\includegraphics[trim={0cm 0.2cm 0cm 0.4cm},clip,width=0.44\textwidth]{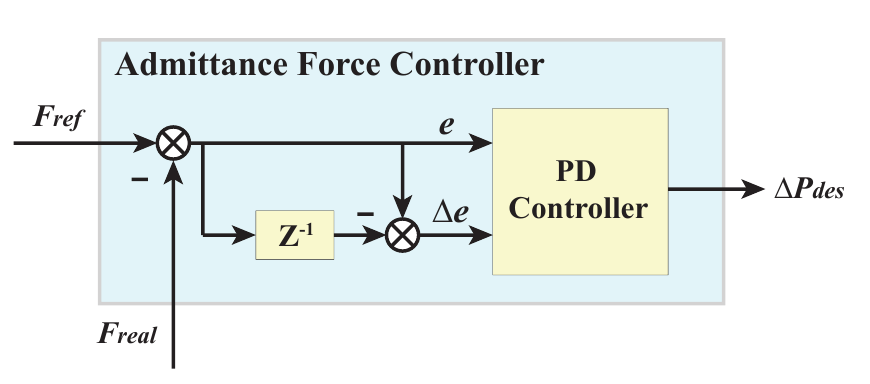}
	\caption{Block diagram of the admittance force controller.}
	\label{fig:force_control}
	\vspace{-2mm}
\end{figure}A regression model from sEMG signals to fingertip forces needs to be built to predict the contact forces. In \cite{Cao2015}, the extreme learning machine (ELM) was applied to predict handgrip force from sEMG where a dynamometer was used to measure the force. However, the dynamometer constrains the fingers' movement, making it difficult to directly transfer the force from humans to the robotic gripper. Therefore, a new data acquisition procedure is needed to gather the gripping forces, which will be used to build the regression model in a supervised manner.

\section{Force-guided High-precision Grasping Control Framework}
\label{sec:iii}
We will overview the force-guided control framework for high-precision grasping and illustrate the realization of an admittance controller for force tracking.

\subsection{Overview}

The overview of the proposed framework is shown in Fig. \ref{fig:control_diag}. Inputs of the system are the multichannel sEMG signals measured by electrodes placed on the forearm. To manipulate the objects, the interaction between the robotic gripper and objects can be modeled as a series-elastic system, in which a change in position (deformation) results in a change of force \cite{zlistabilizer2012}. Hence, an incremental change of force can be mapped to a position command to the gripper. The force control is implemented by an admittance controller that transforms the force reference to a position command for the inner loop \cite{li2015compliance}. We obtain a force-controlled gripper with an inner position loop, where the force reference is predicted from the multichannel sEMG signals by a neural network.

\begin{figure*}[t]
	\centering
	\includegraphics[trim={0cm 0cm 0cm 0cm},clip,width=0.9\textwidth]{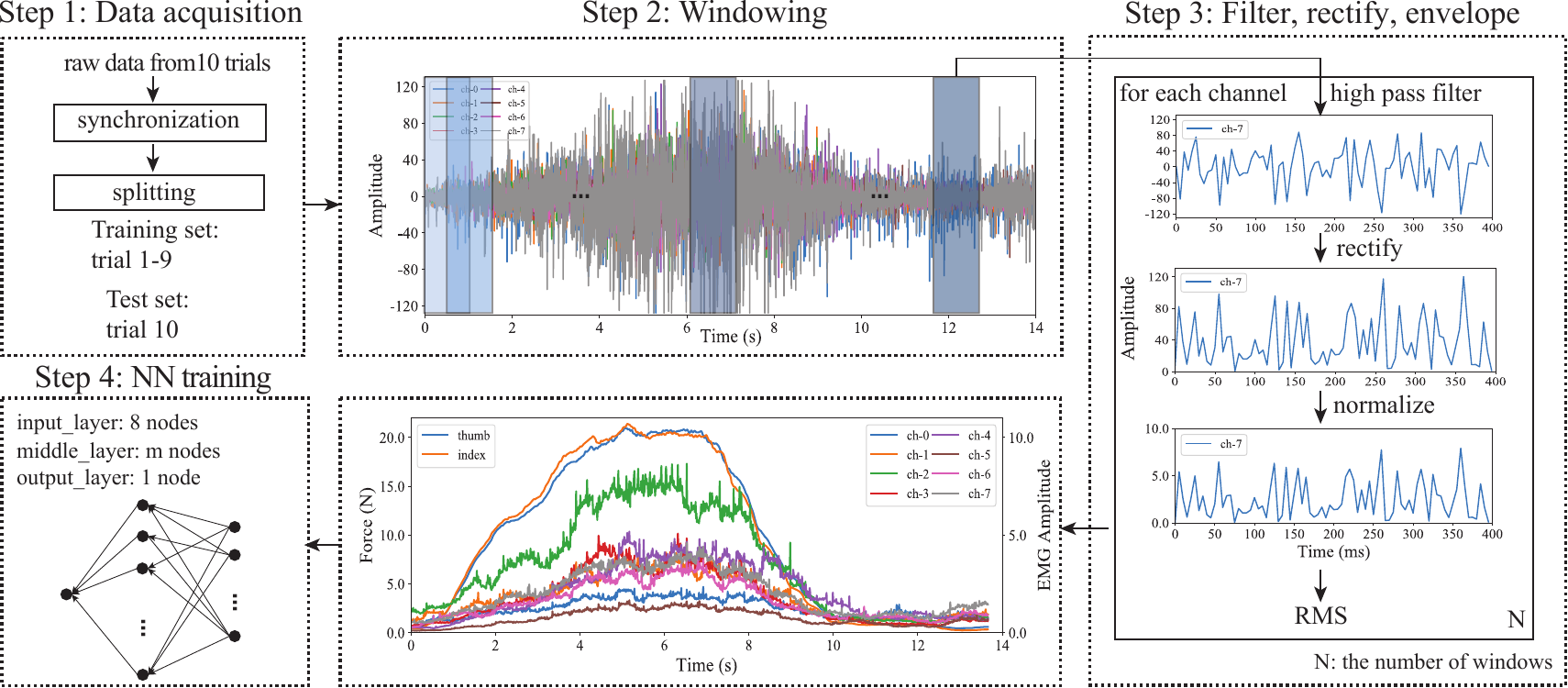}
	\caption{Framework of the sEMG-based gripping force prediction, including sEMG and force data acquisition, sequence windowing, envelop extraction and neural network training.}
	\label{fig:pred_diag}
	\vspace{-5mm}
\end{figure*}

\subsection{Force Regulation via Admittance Control}

By using force measurements, we design a controller that can achieve an outer closed-loop control of the contact forces, as shown in Fig. \ref{fig:force_control}.

The tracking control law of the contact force is formulated as follows:
\begin{equation}
\label{eq:IC}
\begin{aligned}
\Delta{P}_{des}(N) &= K_p \cdot e(N) + K_d \cdot \Delta{e}(N) \\
P_{des}(n) &= P_{des}(N-1) + \Delta{P}_{des}(N)
\end{aligned},
\end{equation}
where $e = F_{ref} - F_{real}$ is the force error and $P_{des}$ denotes the desired position which will be sent to the inner position loop. This principle is similar to the force control using series elastic actuators.

\section{Gripping Force Prediction based on sEMG}
\label{sec:iv}

The sEMG-based gripping force prediction model was built by following the steps shown in Fig. \ref{fig:pred_diag}, including data acquisition, preprocessing, and model training. 

\subsection{Data Acquisition}
To build the prediction model, we gathered raw multichannel sEMG signals as the inputs and forces measured from the thumb and index fingertips as the labels. Because sEMG measures the electric potential generated by multiple activated muscle cells, it is a non-stationary stochastic process. Considering its time-variant feature even for the same task, we gathered  data from 10 trials. Before each trial, the subject was seated in a chair facing the computer monitor with the thumb and index fingertips touching the force sensors, forearm muscles in a rest state, and fingertips not generating any force. During each trial, the subject performed an increasing force exertion as smooth as possible to the maximum force (without reaching muscular fatigue) within 10 s by observing the force curves on the screen.  

\subsection{Maximum Voluntary Contraction}

For EMG normalization, the maximum voluntary contraction (MVC) task was performed via three trials. Each trial involved a gradual increasing process of the contact force. Forces were exerted by the thumb and index fingers to the sensors in the normal direction to the contact surface. The force amplitude went up from baseline to peak in 3-4 s and then was kept for approximately 2 s. If the difference of the peak forces between two trials exceeded $5\%$, a subsequent trial would be performed. Between trials, there was a rest period of 1 min. To get the MVC profile of the sEMG, a 400 ms window centered at the time when the force reached the peak was chosen. For each channel of sEMG signals, the MVC profile was defined as the root mean square (RMS) amplitude within the window, calculated as follows:
\begin{equation}
\label{eq:RMS}
\begin{aligned}
RMS = \sqrt{\frac{1}{N}\sum_{n=1}^{N} x_n^2} ,
\end{aligned}
\end{equation}
where $N$ denotes the number of samples within each sliding window and $x_n$ denotes the $n^{th}$ sample.

\subsection{Preprocessing}

First, multichannel sEMG signals and the force data from the thumb and index fingertips were synchronized. Then, for the raw sEMG signal from each trial, we did segmentation using a sliding window with a 10 ms increment. Raw sEMG signals (Step 2 in Fig. \ref{fig:pred_diag}) contain noise from moving artifacts, baseline, electrocardiogram and power line hum. To eliminate the DC component, each channel of the sEMG signals was first high pass filtered by a Butterworth filter (4th order, cutoff frequency of 5 Hz). Each channel was rectified and then normalized by the MVC profile, as shown in \eqref{eq:rectify_norm}. 

\begin{equation}
\label{eq:rectify_norm}
\begin{aligned}
x_{hp}(t) &= \mathcal{H}(x(t))\\
x_{rect}(t) &= |x(t)| \\
x_{norm}(t) &= \frac{x_{rect}(t)}{x_{mvc}}\\
\end{aligned},
\end{equation}
where $x(t)$ is the $t^{th}$ sample within each processing window, $\mathcal{H}$ denotes the Butterworth high pass filter, $|\cdot|$ is the absolute function, and $x_{mvc}$ is the MVC profile of sEMG.

\begin{figure}[h!]
	\centering
	\includegraphics[trim={0cm 0.3cm 0cm 0.2cm},clip,width=8.2cm]{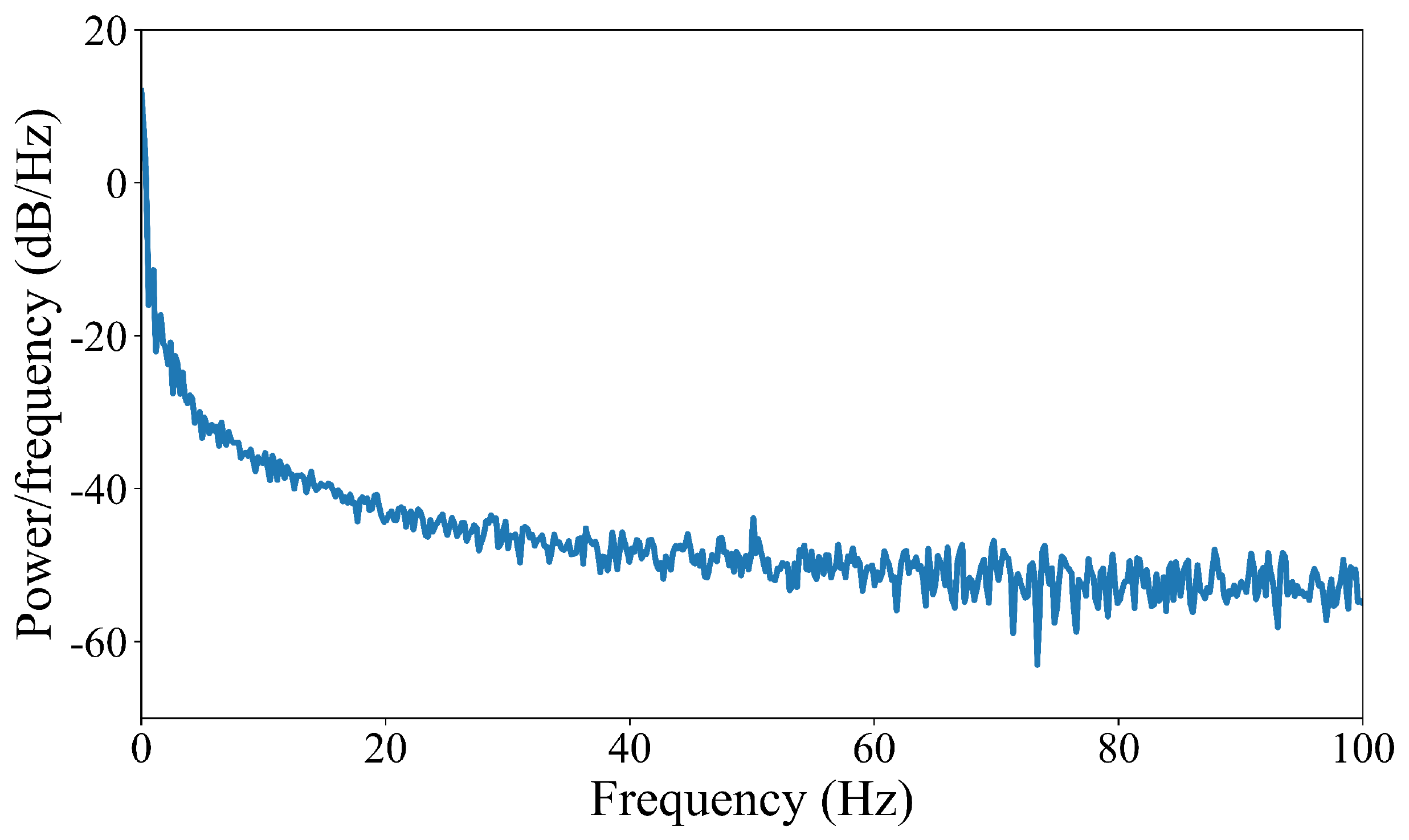}
	\vspace{-1mm}
	\caption{Power Spectral Density of the grasping force using FFT.}
	\label{fig:power_spectrum}	
	\vspace{-4mm}
\end{figure}

The RMS value was considered to reflect the physiological correlation of the motor unit behavior during contraction, which has a quasi- or curvilinear-relationship with the force exerted by a muscle \cite{c4}. Moreover, it can be easily implemented in both digital and analog systems, defined by \eqref{eq:RMS}.

For force data, a Fast Fourier Transform (FFT) was used to analyze the power spectral density (PSD) during the grasping process. The result is shown in Fig. \ref{fig:power_spectrum}, which indicates force signals are low-frequency signals, therefore, we used a low pass filter with the cutoff frequency of 15 Hz to remove the white noise while keeping the most useful information. Moreover, considering EMD between the start of the muscular activation and the force generation at the fingertip, forces at the end of each window were used as the labels corresponding to the preprocessed sEMG signals within the same window.

\subsection{Offline Neural Network Training}
\begin{figure}[h!]
	\centering
	\includegraphics[width=8.2cm]{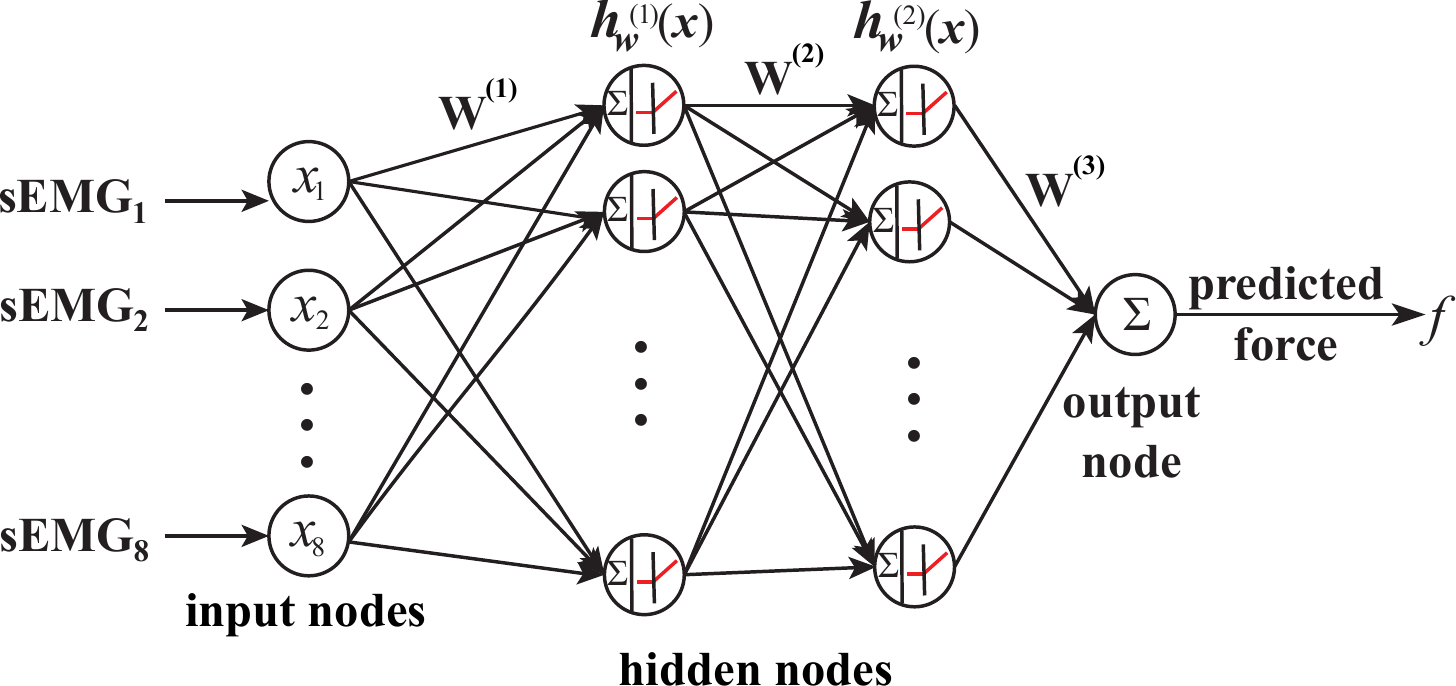}
	\caption{Neural network based force prediction from sEMG.}
	\label{fig:NN}	

\end{figure}

The muscular activity and the gripping force have a nonlinear relationship, therefore, the prediction was modeled as a nonlinear regression problem which can be solved by supervised learning with EMG data as inputs and measured gripping forces as labels. A multi-layer perceptron (MLP) as a feedforward neural network with multiple layers uses backpropagation algorithm to minimize the errors between the prediction and the measured gripping force. In this work, a three-layer neural network using ReLU (rectified linear unit) as the activation function with eight input nodes and one output node was trained in a mini-batch by the ADAM optimizer, the structure of which is illustrated in Fig. \ref{fig:NN}.

The mathematical representation of the neural network is:
\begin{equation}
\label{eq:NN}
\begin{aligned}
h^{(1)}_{j} &= g(\sum_{i=0}w^{(1)}_{ij}x_i), \\
h^{(2)}_{k} &= g(\sum_{j=0}w^{(2)}_{jk}h^{(1)}_j), \\
f &= \sum_{k=0}w^{(3)}_{k}h_k,
\end{aligned}
\end{equation}
where $x_i$ is the $i$ th input node, and $g(\cdot) = max(0,\cdot)$ represents the ReLU activation function.

The loss function is defined as the mean squared error between the predicted and measured force, where L2 
\begin{table}[t]
	\caption{Hyperparameters of the force-guided grasping system} %title of the table
	\centering % centering table
	\begin{tabular}{l l l } % creating eight columns
		\hline\hline %inserting double-line
		Parameters&Search space&Selected Value \\ [0.5ex]
		\hline
		batch size & 32, 64, 128& 64\\ % Entering row contents
		learning rate &0.01, 0.005, 0.001, 0.0001& 0.001 \\
		weight decay & 0.05,0.01,0.005& 0.01\\
		num of hidden nodes & 10,15,20& 15\\
		num of epoch & [0,500] & 50\\
		$K_p$ & --- & 0.1\\
		$K_d$ & --- & 0.002\\[1ex] % [1ex] adds vertical space
		\hline % inserts single-line
	\end{tabular}
	\label{tab:hyper}
	\vspace{-3mm}
\end{table}regularization is used to prevent overfitting:
\begin{equation}
\label{eq:loss}
\begin{aligned}
J &= \dfrac{1}{N}\sum_{n=1}^{N} (f_n - y_n)^2, \\
L &= J(\bm{W};\bm{x},\bm{y}) + \lambda \norm{\bm{W}}^2,
\end{aligned}
\end{equation}
where $N$ is the batch size, $\norm{\cdot}$ is the L2 norm, $\bm{W}$ is the weight matrix, and $\lambda$ is the regularization parameter.

The weights were updated for each batch via backpropagation. To optimize the hyperparameters including batch size, learning rate, weight decay and the number of hidden layer nodes, we run 10-fold cross-validation with grid search on the searching space as shown in  Table \ref{tab:hyper}. Early stopping technique was also used to prevent overfitting the model.

\subsection{Prediction Model Assessment}
The performance of the offline trained neural network is assessed by the coefficient of determination ($R^2$) index which is defined as follows:
\begin{align*} \label{eq:r2}
R^2 &= 1 - \dfrac{\displaystyle\sum_{i=1}^{M}(y_i - \hat{y_i})^2}{\displaystyle\sum_{i=1}^{M}(y_i - \bar{y})^2},
\end{align*}
where $y_i$ and $\hat{y_i}$ denote measured and predicted gripping force values for the $i^{th}$ sample in the data set, and $\bar{y}$ is the mean of the measured force over all the samples.

\subsection{Online Gripping Force Prediction}
Using the offline trained neural network, the force reference was predicted in real time by Algorithm \ref{al:Pred}, where FIFO represents the first-in-first-out buffer.  Considering the frequency feature of the human's gripping force, we set the control frequency of the outer force loop to 20 Hz, which was adjusted by the size of the EMG and force data buffer.

\begin{algorithm}[t]
	\caption{Online Force Prediction}
	\label{al:Pred}
	\begin{algorithmic}[1]
		\REQUIRE ~~ \\
		Multichannel EMG signals
		\ENSURE ~~ \\
		Gripping force reference
		\STATE Initialize an empty EMG FIFO buffer.
		\STATE Initialize an empty Force buffer.
		\IF{EMG buffer not full}
		\STATE Append the EMG data to the head of the EMG buffer.
		\ELSE
		\STATE Preprocess the data in the EMG buffer and predict the force using the trained NN model.
		\IF{Force buffer is not full}
		\STATE Append the predicted force to the head of the Force buffer.
		\ELSE
		\STATE Calculate the mean of the force value in the Force buffer as the gripping force reference.
		\STATE Empty the Force buffer.
		\ENDIF
		\STATE Discard the EMG data at the buffer tail.		
		\ENDIF	
	\end{algorithmic}
	
\end{algorithm}

\section{Experimental Setup}
\label{sec:experiment setup}

Before conducting the grasping tasks of fragile and deformable objects, two parts -- sensors for data acquisition and the robotic gripper for grasping -- need to be set up.

\subsection{Setup of Data Acquisition}
Thalmic Myo armband was a myoelectric device with eight embedded electrodes for gathering sEMG data of a group of forearm muscles. Two OptoForce OMD-30-SE-100N 3D force sensors fixed on a 3D printed cube were used to measure the magnitude and direction of the force applied by the fingertips. The subject applied forces at the poles of two semi-spherical sensors in the normal direction to the contact surface. The monitor displaying the force amplitude helped the subject to generate forces as smoothly as possible (Fig. \ref{fig:data_acq_setup}).

\subsection{Setup of Grasping Tasks}
\begin{figure}[t]
	\centering
	\subfigure[]{
		\label{fig:data_acq_setup}
		\includegraphics[trim={0cm 0cm 0cm 0cm},clip,height=4.0cm]{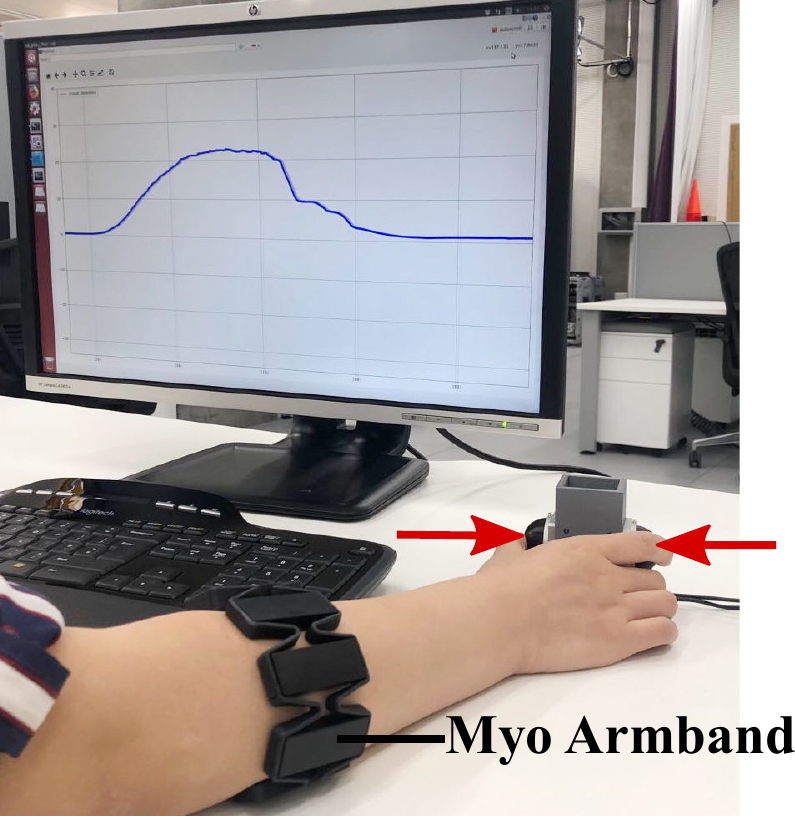}}
	\hspace{1mm}
	\subfigure[]{
		\label{fig:tf}
		\includegraphics[trim={0cm 0cm 0cm 0cm},clip,height=4.0cm]{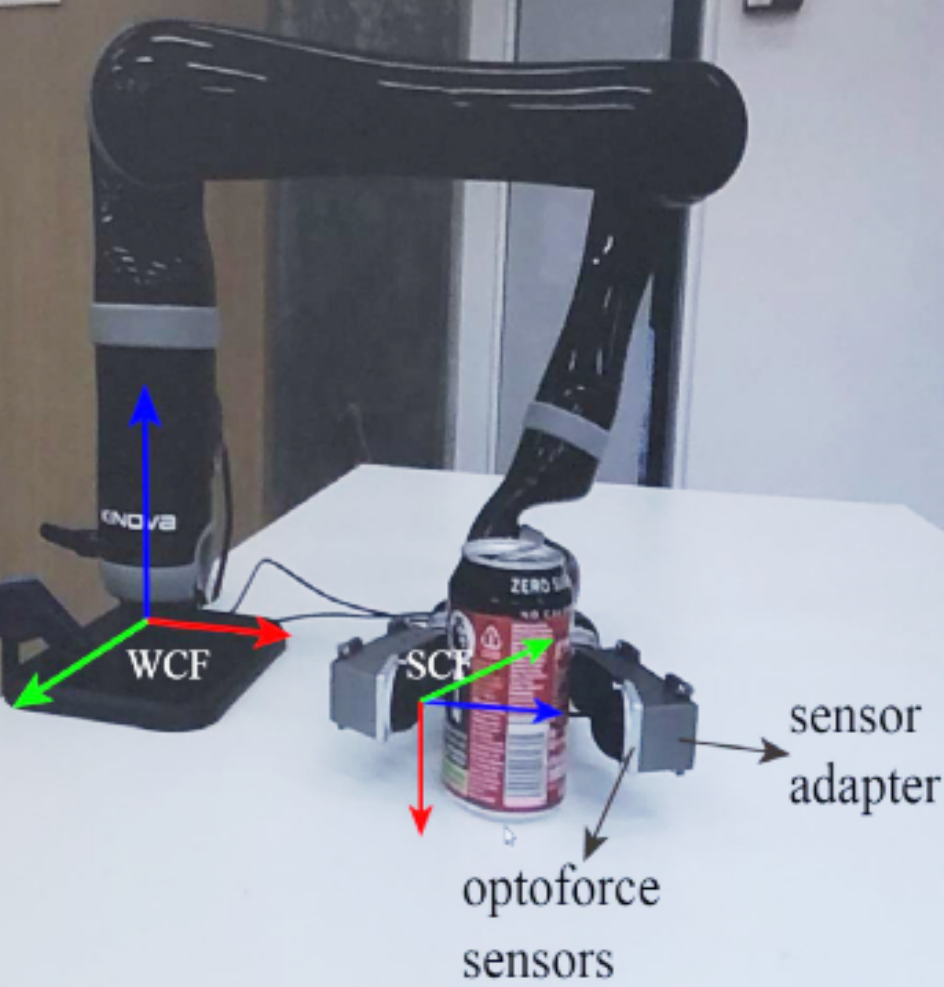}}
	\caption{The experimental setup of data acquisition for building sEMG-based force prediction model and the coordinate frames. (a) the data acquisition setup; (b) the world and sensor coordinate frame (WCF, SCF).}
	
\end{figure}

We used a 6-DOF Kinova Arm with a two-finger gripper (KG-2), of which each finger is composed of a proximal and distal phalanx. The opening and closing movements of the two fingers are driven by two linear actuators, one for each finger, which controls the movement of the distal phalanx. To measure the magnitude and direction of the contact force between the fingertip and the objects, two OptoForce OMD-30-SE-100N 3D force sensors were fixed on the grippers by the 3-D printed adapters, which were specially designed so that the z-axis was horizontal and the x-axis pointed downwards when the gripper was at the posture shown in Fig. \ref{fig:tf}. 

\begin{figure}[t]
	\centering
	\includegraphics[trim={0cm 0.2cm 0cm 0cm},clip,width=8.5cm]{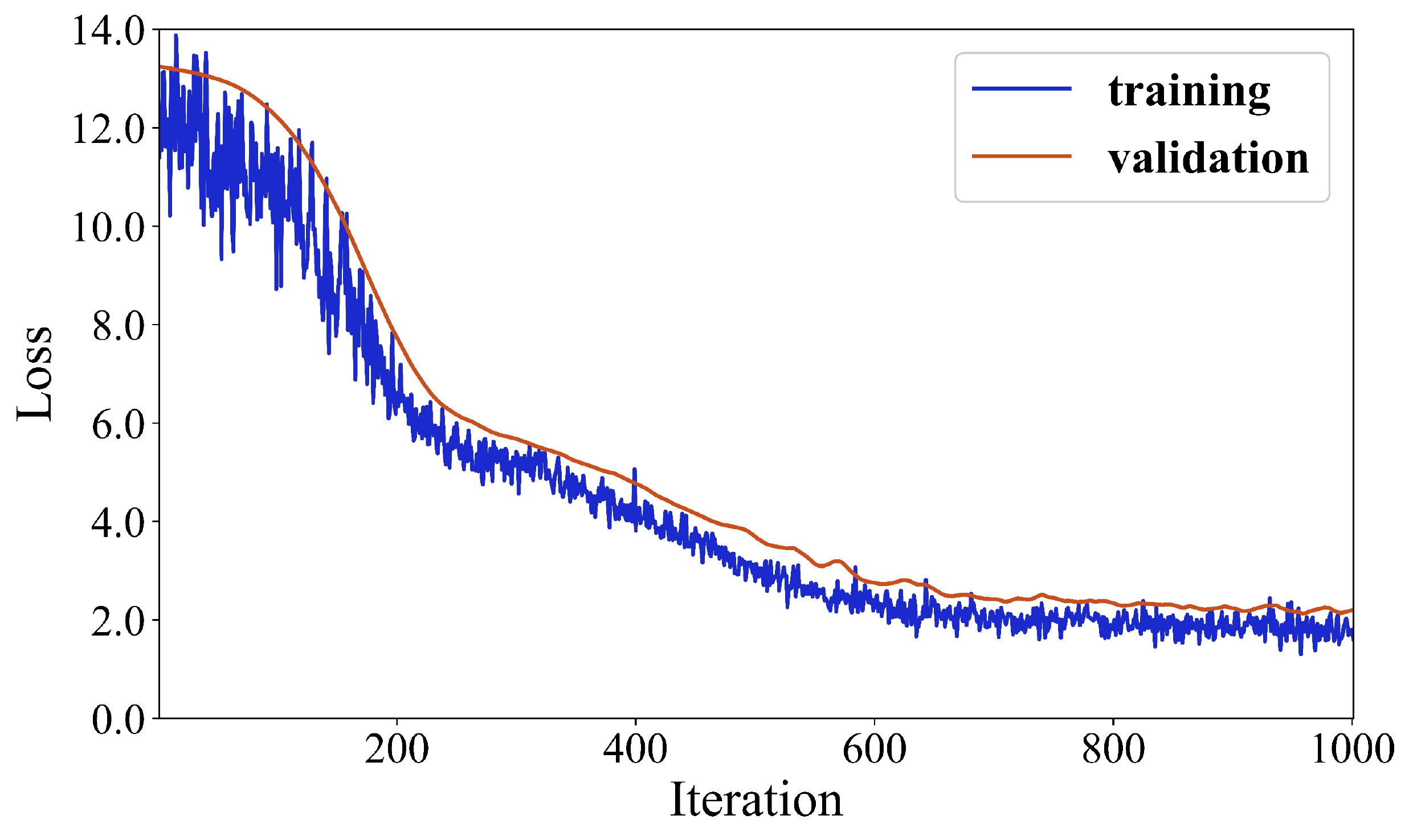}
	\vspace{-3mm}
	\caption{Loss curve during the training of sEMG-based gripping force prediction model.}
	\label{fig:loss}
	\vspace{-4mm}
\end{figure}
In this paper, we discuss the force in two coordinate frames. At the human side, since the prediction model was trained in the world coordinate frame (WCF), the gripping force was predicted in WCF. At the robot side, the contact force was measured in the sensor coordinate frame (SCF). We simplified the grasping tasks by placing each object at a fixed place on the table and hard-coding the reaching to the pre-grasp position. To address the mismatch of the contact points between the human and robot side during grasping, the force vectors measured in SCF were transformed to WCF.

At the human side, the Myo armband was worn on the forearm at the same position as that in the training session. The image of the Kinova gripper was transmitted to the human side in real time for visual feedback.

\section{Experimental Validation of the Force-guided High-precision Grasping Control Framework}
\label{sec:results}
In this section, the results of both the sEMG-based force prediction model and the grasping tasks of fragile and deformable objects will be presented and analyzed.

\subsection{Prediction Model}
\begin{figure}[t]
	\centering
	\includegraphics[trim={0cm 0.2cm 0cm 0cm},clip,width=8.5cm]{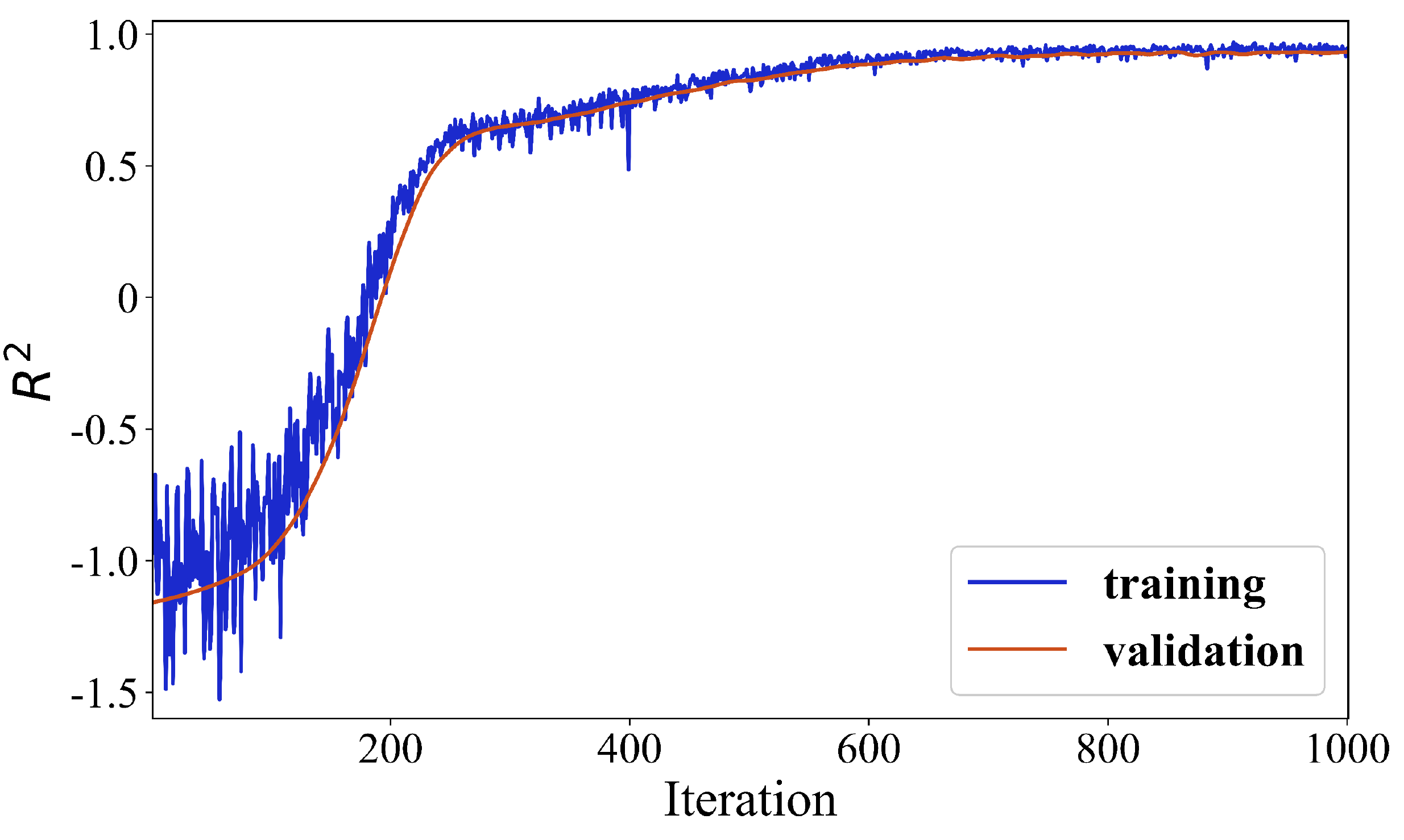}
	\vspace{-1mm}
	\caption{Curve of the coefficient of determination ($R^2$) during the training of sEMG-based gripping force prediction model. $R^2$ reached $0.95$ both on the training and validation set after 700 iterations.}
	\label{fig:r2}
	\vspace{-4mm}
\end{figure}

To evaluate the performance of the neural network, the data set was split into a training ($\#1 - \#7$), a validation ($\#8$, $\#9$) and a test set ($\#10$), as illustrated in Step 4 (Fig. \ref{fig:pred_diag}). The loss and $R^2$ curves during the training process with the optimized hyperparameters (shown in Table \ref{tab:hyper}) are shown in Fig. \ref{fig:loss} and Fig. \ref{fig:r2}.

The root mean squared error (RMSE) between the labels and predicted values was calculated after each iteration on the training and validation sets, shown in Fig. \ref{fig:loss}.

\begin{figure}[t]
	\centering
	\includegraphics[trim={0cm 0.4cm 0cm 0cm},clip,width=8.5cm]{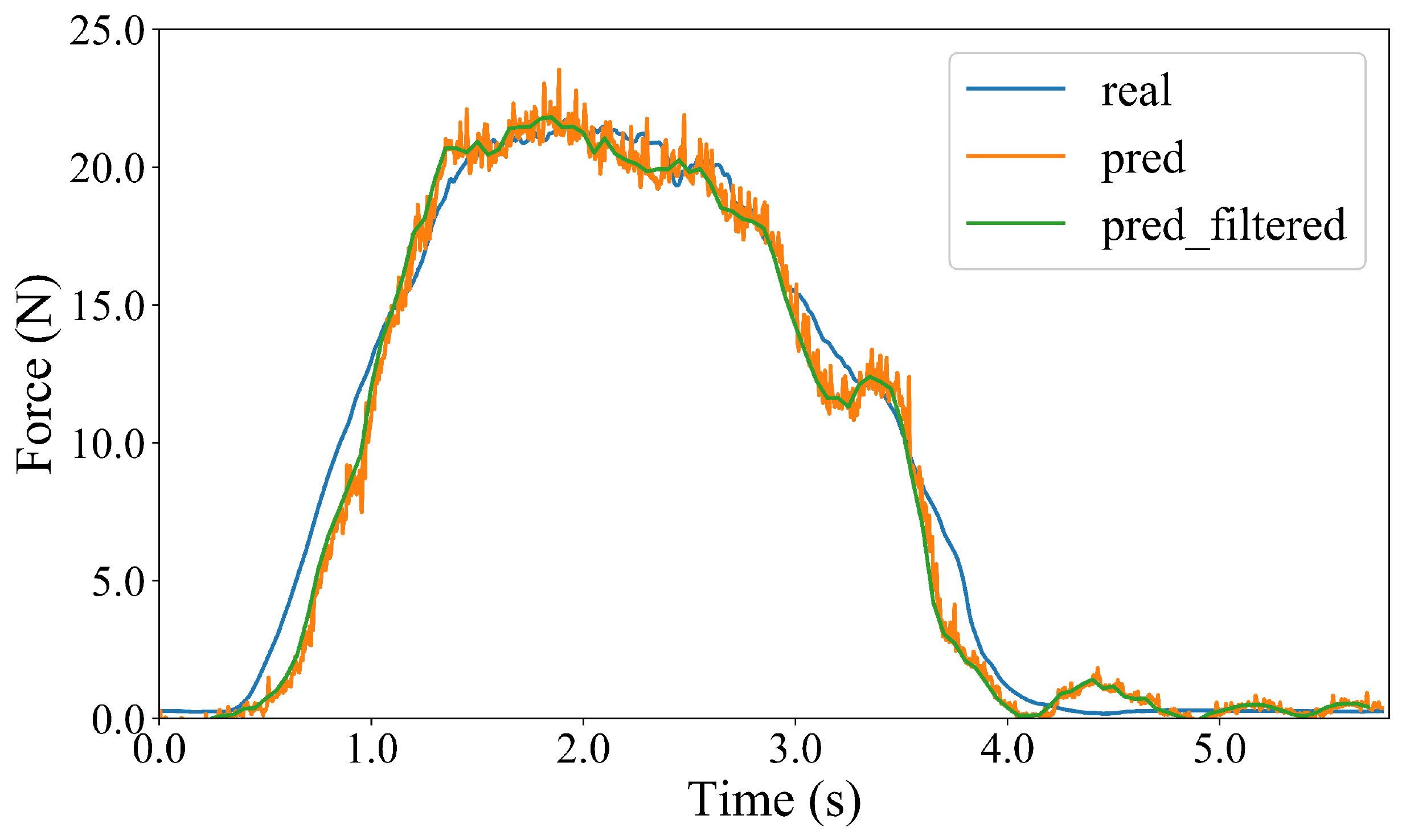}
	\vspace{-1mm}
	\caption{Comparison of raw/filtered predicted force against the ground truth.}
	\label{fig:groundtruth}
	\vspace{-2mm}
\end{figure}

The $R^2$ index on both data sets went up soon after the training started and fluctuated around $0.95$ after 700 iterations. $R^2$ on the validation set did not have a salient difference from that on the training set, indicating that the proposed prediction model has a good generalization performance. The test set was then used to further evaluate the performance of the prediction model (shown in Fig. \ref{fig:groundtruth}). By applying a moving average window to smooth the raw predicted force, $R^2$ of the prediction model had an increase from $0.9795$ to $0.9820$.

\subsection{Grasping Tasks with the Proposed Method}

\begin{figure*}
	\centering
	
	\subfigure[Fruit pepper.]{
		\label{fig:pepper}
		\includegraphics[trim={0cm 0cm 0cm 2.5cm},clip,height=3.5cm]{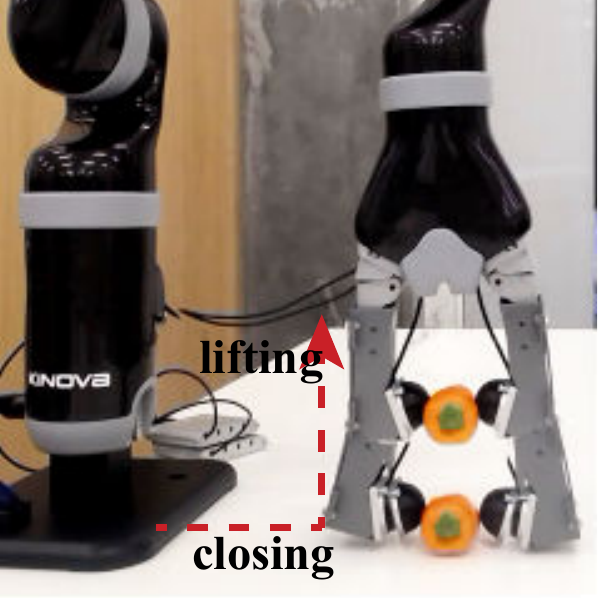}}  \ \ \
	\subfigure[Ripe tomato.]{
		\label{fig:tomato}
		\includegraphics[trim={0.2cm 0cm 0cm 2.5cm},clip,height=3.5cm]{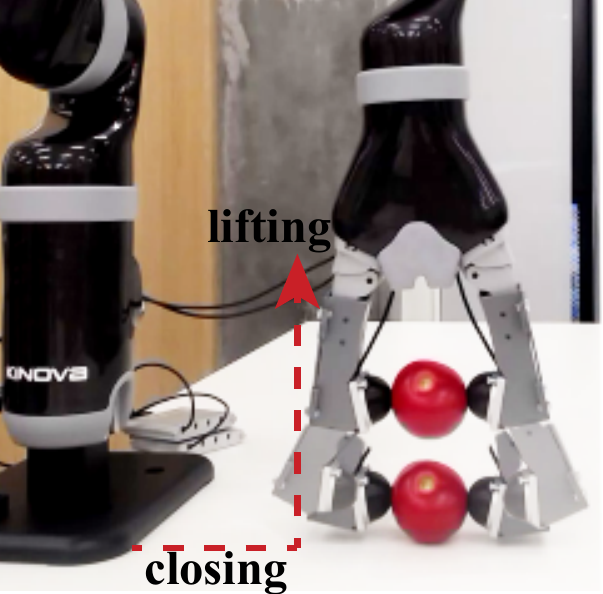}}\ \ \
	\subfigure[Full plastic water bottle.]{
		\label{fig:bottle}
		\includegraphics[trim={0cm 0cm 0cm 0cm},clip,height=3.5cm]{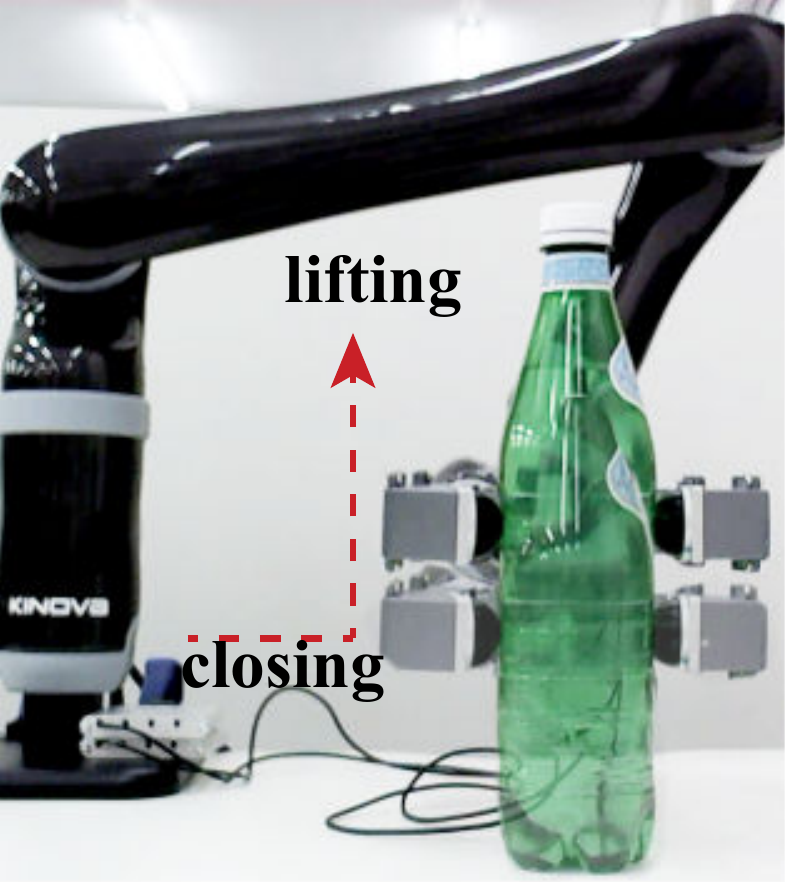}} \ \
	\subfigure[Empty aluminum can.]{
		\label{fig:can}
		\includegraphics[trim={0cm 0cm 0cm 0cm},clip,height=3.5cm]{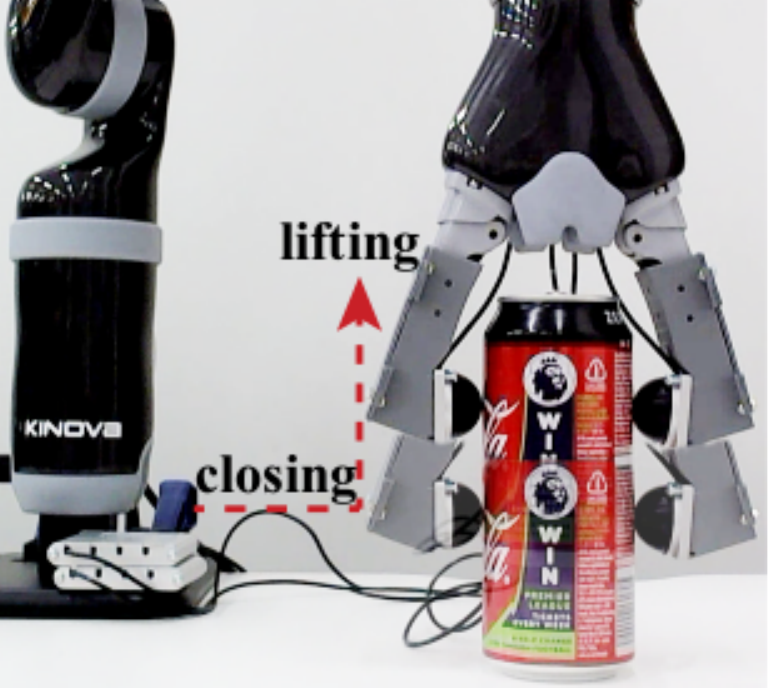}}
	\subfigure[Ripe strawberry.]{
		\label{fig:strawberry}
		\includegraphics[trim={0cm 0cm 0cm 0cm},clip,height=3.5cm]{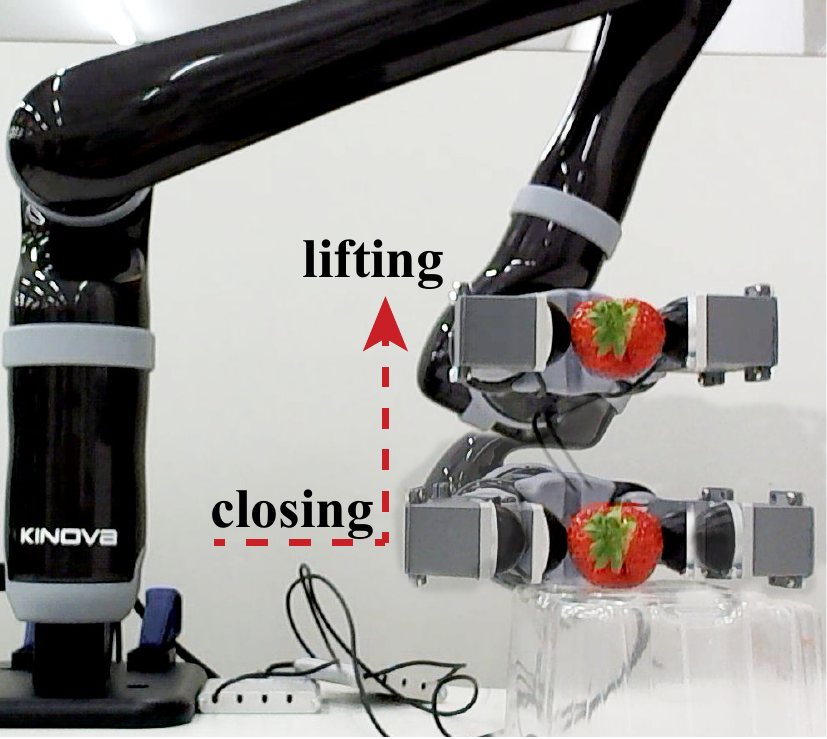}}
	\subfigure[Half egg shell (very delicate).]{
		\label{fig:eggshell}
		\includegraphics[trim={0cm 0cm 0cm 0cm},clip,height=3.5cm]{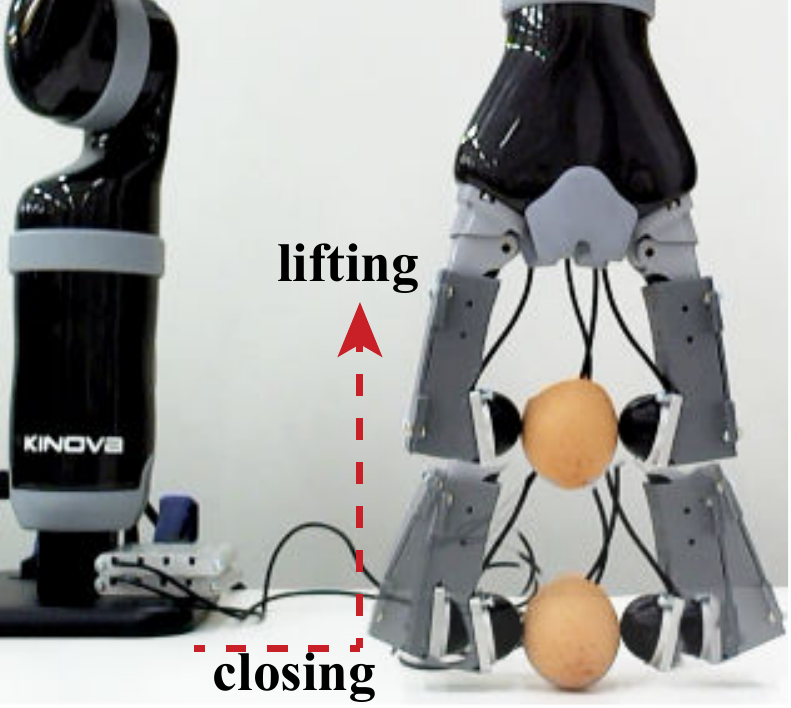}}
	\subfigure[Thin glass (0.5 mm).]{
		\label{fig:glass_slice}
		\includegraphics[trim={0cm 0cm 0cm 0cm},clip,height=3.5cm]{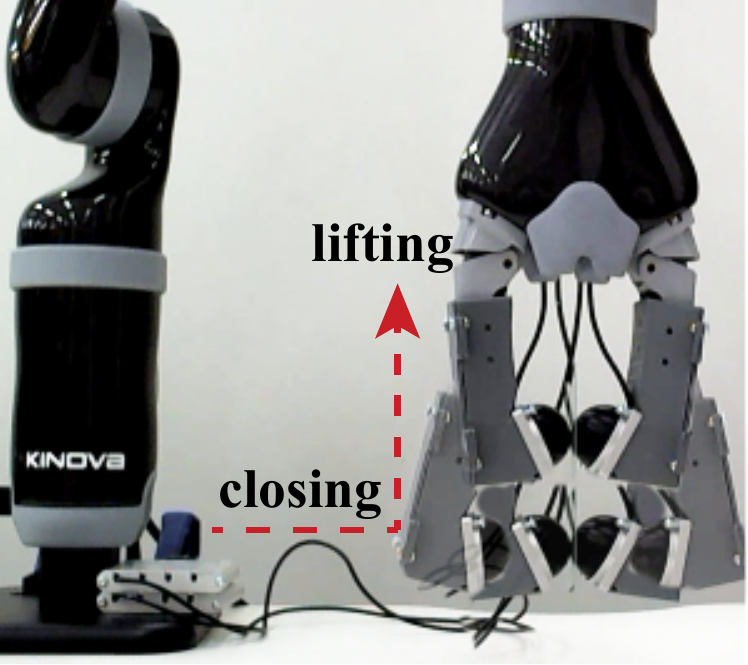}}
	\subfigure[Wine glass.]{
		\label{fig:glass}
		\includegraphics[trim={1.3cm 0cm 0.7cm 0cm},clip,height=3.5cm]{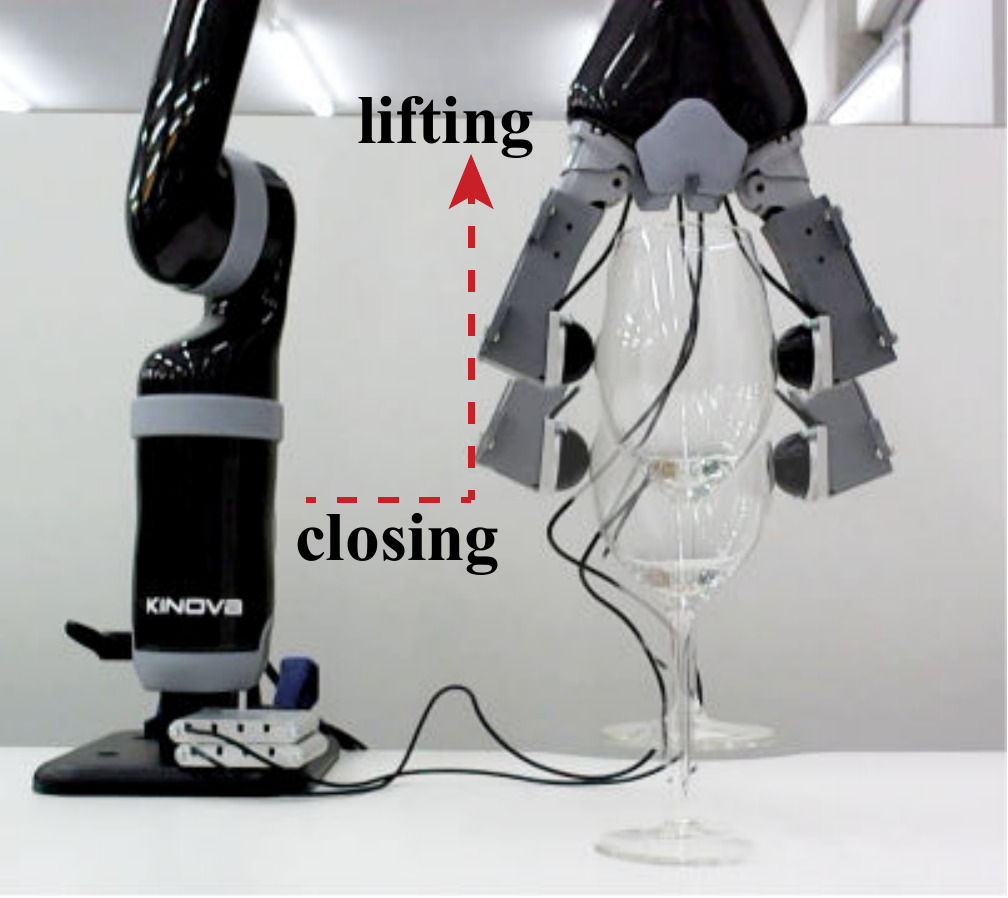}}
	\caption{Grasping tasks on eight representative deformable and fragile objects from daily life. The overlaying pictures show the closing gripper using active force regulation from an operator, followed by the lifting motion to indicate successful grasping. Please see more details of the grasping process and more trials of objects in the accompanying video.}	
	\label{fig:tasks}
	\vspace{-2mm}
\end{figure*}

To validate the performance of the proposed force-guided grasping framework, three deformable (fruit pepper, full plastic bottle, empty aluminum can) and five fragile (ripe tomato, ripe strawberry, half eggshell, 0.5 mm thin glass slice, wine glass) objects were tested. With the experimental setup, the robotic gripper successfully grasped all objects without deforming or breaking them by tracking the guided force predicted from sEMG of the subject when executing the same tasks on the other side. The success of these tasks was verified by lifting the object, which occurred once the difference between the predicted and measured force was less than a threshold of 0.01 N (Fig. \ref{fig:tasks}). The predicted force, the measured force and its percentage of gripper's maximum force (20 N) at the end of the grasping phase, are shown in Table \ref{tab:force}. Without the predicted force as the reference, it will take a long time to learn the correct contact force, and meanwhile causing many inevitable failures, either deforming or damaging the objects. On the other hand, using our proposed method, the operator had one-shot success for all the tasks within the test sets of studied objects, which was attributed to the guided force and prior human grasping synergies of regulating motion and force profiles. 
 
\section{Discussion}
\label{sec:discuss}
The success of the force-guided grasping control framework is attributed to the high accuracy of the sEMG-based force prediction model and the admittance controller for force tracking. We applied an incremental control law on the tracking error in the inner position loop, which can have improved control bandwidth using a torque-controlled robotic hand. In addition, the uncertainty during the grasping can be compensated by the human operator from the visual feedback via video streaming. In this regard, the framework fulfills the goal of utilizing the human capability of motion-force synergies to adjust the contact forces, which can be further used to collect data of human grasping demonstrations as future work.
Our work also has some limitations: in order to generate suitable contact forces to guide the robotic grippers, humans need to have physical access to the duplicated mock-up objects to sense the weight and texture. Humans have the ability to learn and adjust the gripping force by manipulating virtual objects. The development of virtual reality techniques can render the collision between human fingers and objects, through which humans can learn contact forces with haptic feedback. Therefore, the necessity of the mock-up objects is eliminated. The guided-force reference can be obtained without causing any damage to real objects. Moreover, our method can be applied to build large data sets of the contact force profile for the objects where the force sensors cannot be easily installed. 

\begin{table}[t]
	\caption{Predicted force, real force and its percentage of the gripper's maximum force.} %title of the table
	\centering % centering table
	\begin{tabular}{l l l r} % creating three columns
		\hline\hline %inserting double-line
		Objects&Predicted (N)&Real (N)&Max ($\%$)\\ [0.5ex]
		\hline
		Fruit pepper &1.140& 1.133&5.67\\ % Entering row contents
		Ripe tomato &0.239& 0.231 &1.16\\
		Wine glass & 5.932& 5.922&29.61\\
		Aluminum can & 0.216& 0.214&1.07\\
		Thin glass (0.5mm) & 0.826 & 0.826&4.13\\
		Strawberry & 0.039 & 0.040&0.20\\
		Full plastic bottle& 6.770 & 6.774&33.87\\
		Egg shell& 0.010 & 0.008&0.04\\[1ex] % [1ex] adds vertical space
		\hline % inserts single-line
	\end{tabular}
	\label{tab:force}
	\vspace{-2mm}
\end{table}

To obtain the contact force feedback, we integrated the OptoForce sensors on the grippers of the Kinova Arm platform through 3-D printed adapters, which is not necessary for grippers with force measurements. There are also commercial substitutes for Thalmic Myo armband (Delsys, Ottobock, etc). Therefore, the proposed method is not specifically bounded to the hardware of this study and can be transferred to other robotic platforms as well.

\section{Conclusion and Future Work}
\label{sec:conclu}
In this paper, we proposed a force-guided framework for fragile and deformable objects via a myoelectric interface. The proposed method predicts the guiding forces from sEMG on the forearms of human demonstrators using an artificial neural network, and does not require any prior knowledge of objects regarding shape, weight and surface friction. The force commands were then tracked by an admittance controller. The performance was validated by successfully grasping more than eight fragile and deformable objects. Under the teleoperation framework, our method utilized humans’ experience of grasping through kinesthetic feedback and visual feedback.

For future work, the proposed framework can be used as an equivalent perception-action interface between the human operator and the robot for realizing autonomous grasping using the machine learning approach. It can provide a large amount of useful human demonstration data to develop data-efficient learning via such human-robot apprenticeship.

\section{Acknowledgment}
This work was supported in part by the National Natural Science Foundation of China (61876054), in part by the China
Scholarship Council, in part by the EPSRC CDT in Robotics and Autonomous Systems (EP/L016834/1), and in part by the EPSRC Future
AI and Robotics for Space (EP/R026092/1).

\bibliographystyle{IEEEtran}
%\bibliography{Ref}

\end{document}